\documentclass{article}

\usepackage{microtype}
\usepackage{graphicx}
\usepackage{subfigure}
\usepackage{booktabs} 

\usepackage{hyperref}

\usepackage[accepted]{icml2024}

\usepackage{amsmath}
\usepackage{amssymb}
\usepackage{mathtools}
\usepackage{amsthm}
\usepackage{enumitem}
\usepackage{multirow}
\usepackage{url}

\usepackage[capitalize,noabbrev]{cleveref}

\theoremstyle{plain}

\theoremstyle{definition}

\theoremstyle{remark}

\usepackage[textsize=tiny]{todonotes}

\icmltitlerunning{Position: Foundation Agents as the Paradigm Shift for Decision Making}

\begin{document}

\twocolumn[
\icmltitle{Position: Foundation Agents as the Paradigm Shift for Decision Making}




\begin{icmlauthorlist}
\icmlauthor{Xiaoqian Liu}{sch1,yyy}
\icmlauthor{Xingzhou Lou}{sch2,yyy}
\icmlauthor{Jianbin Jiao}{sch3}
\icmlauthor{Junge Zhang}{yyy}
\end{icmlauthorlist}

\icmlaffiliation{sch1}{School of Integrated Circuits, University of Chinese Academy of Sciences, Beijing, China}
\icmlaffiliation{sch2}{School of Artificial Intelligence, University of Chinese Academy of Sciences, Beijing, China}
\icmlaffiliation{sch3}{School of Electronic, Electrical and Communication Engineering, University of Chinese Academy of Sciences, Beijing, China}     
\icmlaffiliation{yyy}{Institute of Automation, Chinese Academy of Sciences, Beijing, China}

\icmlcorrespondingauthor{Junge Zhang}{jgzhang@nlpr.ia.ac.cn}

\icmlkeywords{Foundation Agent, Offline Reinforcement or Imitation Learning, Self-supervised Pretraining, Alignment with LLMs, Sequential Decision-Making}

\vskip 0.3in
]



\printAffiliationsAndNotice{}  

\begin{abstract}

Decision making demands intricate interplay between perception, memory, and reasoning to discern optimal policies.
Conventional approaches to decision making face challenges related to low sample efficiency and poor generalization. 
In contrast, foundation models in language and vision have showcased rapid adaptation to diverse new tasks. 
Therefore, we advocate for the construction of foundation agents as a transformative shift in the learning paradigm of agents. 
This proposal is underpinned by the formulation of foundation agents with their fundamental characteristics and challenges motivated by the success of large language models (LLMs). 
Moreover, we specify the roadmap of foundation agents from large interactive data collection or generation, to self-supervised pretraining and adaptation, and knowledge and value alignment with LLMs.
Lastly, we pinpoint critical research questions derived from the formulation and delineate trends for foundation agents supported by real-world use cases, addressing both technical and theoretical aspects to propel the field towards a more comprehensive and impactful future.

\end{abstract}

\section{Introduction}

\begin{figure*}[ht]
    \centering
    \includegraphics[scale=0.5]{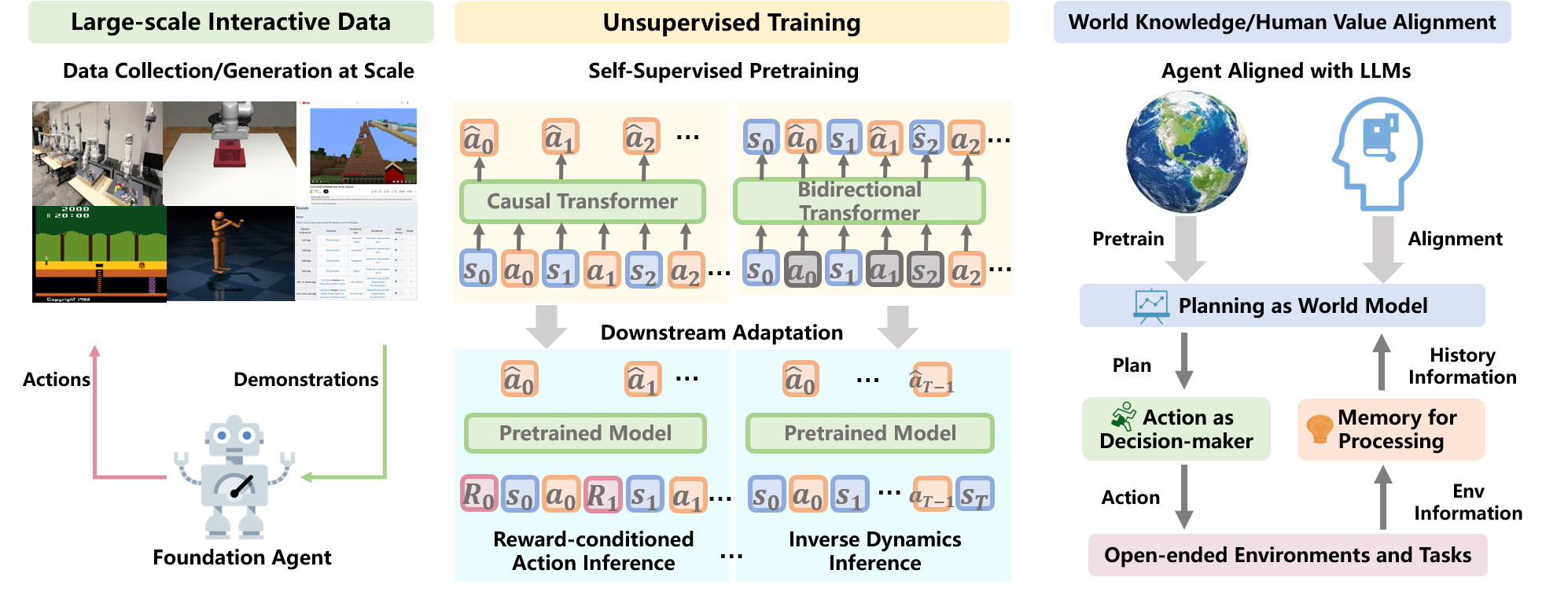}
    \caption{Roadmap of foundation agents. \textbf{Left:} Large interactive data collection or generation scales up foundation agents in open-ended physical and virtual worlds.
    \textbf{Middle:} Self-supervised pretraining leverages the flexible and robust architecture of Transformer \cite{vaswani2017attention} based on autoregressive or masked (in gray) modeling. During adaptation, the pretrained model play various roles in decision making, such as policy initialization \cite{meng2021offline,yang2021representation} and dynamics model \cite{wu2023masked,brandfonbrener2023inverse}.
    Distinct colors denote different variables within trajectories.
    \textbf{Right:} LLMs as a part of foundation agents composed of planning, memory and action modules and act as world models, information processors, and decision-makers, respectively.
    }
    \label{fig:perspectives}
\end{figure*}

Human life is all about making decisions. 
Intelligent agents have been developed to help humans with real-world decision making, such as traffic control \cite{liang2019deep}, energy management \cite{nakabi2021deep}, and drug discovery \cite{zhou2019optimization}. 
Prevalent paradigms to train agents include reinforcement learning (RL), imitation learning (IL), planning and search as well as optimal control. 
Recent advances in these algorithms have achieved superhuman performance in mastering the game of Go \cite{silver2017mastering}, playing video games \cite{schwarzer2023bigger}, and robotic locomotion and manipulation \cite{brohan2022rt}. 
However, traditional approaches to decision making exhibit limited sample efficiency and poor generalization. 
For instance, expert systems heavily rely on human knowledge and manual crafting, and conventional RL methods necessitate agent training from scratch for each task.

In contrast, foundation models~\cite{bommasani2021opportunities} in language and vision achieve rapid adaptation to a wide variety of tasks with minimal fine-tuning or prompting.
These models, pretrained on vast and diverse datasets, demonstrate unprecedented capabilities in understanding and generating text~\cite{brown2020language,gpt4}, image~\cite{dosovitskiy2020image,bai2023sequential} and some multi-modalities~\cite{reed2022generalist,lu2023unified}. 
Therefore, this position paper argues that \textbf{foundation agents as generally capable agents across physical and virtual worlds, will be the paradigm shift for decision making}, akin to LLMs as general-purpose language models to solve linguistic and knowledge-based tasks.

To arrive at this position, we first identify three fundamental characteristics of foundation agents: (1) \emph{a unified representation} of variables involved in decision process, including state-action spaces, feedback signals (e.g., rewards or goals) and environment dynamics, (2) \emph{a unified policy interface} across tasks and domains from robotics and game play to healthcare and beyond, and (3) \emph{interactive decision-making} in physical and virtual worlds by reasoning about behaviors, handling environment stochasticity and uncertainty, and potentially navigating competitive or cooperative multi-agent scenarios.
These characteristics constitute the uniqueness and challenges for foundation agents\footnote{Distinctions from task-specific agents and recent LLM-based agents can be found in Appendix~\ref{append:distinction}.}, empowering them  with multi-modality perception, multi-task and cross-domain adaptation as well as few- or zero-shot generalization. Particularly, foundation agents exhibit enhanced credit assignment and planning in scenarios requiring long-horizon reasoning \cite{wei2022chain,yao2022react,yao2023tree} and involving sparse rewards or partial observability \cite{meng2021offline,guo2023suspicion}. Such unique set of characteristics and capabilities enable foundation agents to improve sample efficiency and generalization, showing their versatility in navigating diverse contexts of decision making.

With the formulation of foundation agents, we specify their roadmap in Figure~\ref{fig:perspectives}. Firstly, \emph{large-scale interactive data} can be collected from the internet (e.g., YouTube videos, tutorials, audios, etc) and physical environments, or generated through real-world simulators~\cite{yang2023learning,sora,bruce2024genie} for supervision and scale-up.
Secondly, \emph{pretrain in an unsupervised manner} from large, unlabeled, and probably suboptimal interactive data for decision-related knowledge representation learning and downstream adaptation with knowledge reasoning. 
Thirdly, align with LLMs to integrate \emph{world knowledge and human values} into foundation agents for reasoning, generalization, and interpretability.
This roadmap is motivated by three key ingredients we observe for the success of LLMs: (1) leveraging internet-scale text data to absorb broad knowledge described by languages, (2) self-supervised pretraining to learn unified text representation~\cite{devlin2018bert} and task interface~\cite{2020t5} through generative modeling, and (3) safety and preference alignment~\cite{ouyang2022training,rafailov2024direct} to fulfill user commands.

For foundation agents, however, it is challenging to reproduce the roadmap of LLMs.
Firstly, the broad information in physical and virtual worlds are low-level details instead of high-level abstractions expressed by languages, posing challenges to a unified representation. Therefore, we discuss the morphology of foundation agents (e.g, a unified or compositional foundation model) for structuring diverse environments and tasks in (\S6.1).
Secondly, the large domain gap between different decision making scenarios makes it difficult to develop a unified policy interface, while linguistic tasks can be simply unified by text generation. We thus posit that theoretical guarantees for policy optimization with foundation agents may address this issue in (\S6.2).
Thirdly, while language and vision models focus on understanding and generating content, foundation agents are involved in dynamic process of choosing optimal actions based on complex environment information. Open-ended tasks and environments are then another critical research question for foundation agents as discussed in (\S6.3).
Finally, we showcase some examples supported by recent work and a case study towards building a potential foundation agent in real-world decision making.

\section{Preliminaries}
\subsection{Decision Making Formalism}

Decision making refers to the process of making a series of decisions to achieve a goal in continuous time, based on previous actions and observations while considering possible future states and rewards. The process can be simplified as a markov decision process (MDP) $M=<S,A,T,R,\gamma>$,
where $S$ is the state space, $A$ the action space, $T$ the dynamics
transition function $T:S\times A\times S\to [0,1)$, $R$ the reward
function $R:S\times A\times S\to \mathbb{R}$, and $\gamma \in [0, 1)$ is a discount
factor for calculating discounted cumulative rewards. 
When the underlying state is not accessible (e.g, a video game), the process can be modified as a partially observable MDP $M=<S,A,T,R,\mathcal{O},E>$, where $\mathcal{O}$
is the observation space, and $E(o|s)$ denotes the observation emission function. Solutions to MDP typically involve RL to learn an optimal policy that maximizes the expected discounted cumulative rewards $\mathbb{E}[\sum_{t=0}^{T}\gamma^tr_{t+1}]$ through trial and error within an environment, or IL to learn an expert's policy from expert demonstrations in a supervised learning manner.
However, real-world decision making is more complex and not limited to MDP, such as stock market dynamics, epidemiological modeling, and sequential games with incomplete information (e.g, Poker and Bridge).

\subsection{Self-Supervised Learning for RL}
To improve sample efficiency and generalization, conventional self-supervised learning for RL involves representation learning of separate RL components, such as action representation \cite{pmlr-v162-gu22b}, reward representation \cite{ma2022vip}, policy representation \cite{tang2022inputting}, and environment or task representation \cite{Sang2022PAnDRFA}. 
Recently, by formulating RL as a sequence modeling problem \cite{chen2021decision,Janner2021OfflineRL}, the representation of various RL components can be simultaneously learned via trajectory optimization \cite{lee2022multi,liu2022masked,carroll2022uni,sun2023smart,wu2023masked,boige2023pasta}.
Such a formulation draws upon the simplicity and scalability of Transformer \cite{vaswani2017attention}, and benefits representation learning of knowledge concerning different aspects of decision process.

\subsection{Large Language Models and Agents}
Large language models \cite{radford2019language,brown2020language,ouyang2022training} are pretrained on internet text and fine-tuned on human instructions and preferences. 
Through this process, LLMs acquire extensive world knowledge and are aligned with human values \cite{gpt4,claude2,rafailov2024direct,touvron2023llama}. 
State-of-the-art LLMs demonstrate convincing capabilities in not only natural language processing \cite{volske2017tl,singhal2022large,jiao2023chatgpt}, but also tasks requiring strong reasoning abilities like coding \cite{nijkamp2022codegen} and text-based games \cite{shridhar2020alfworld,liu2023agentbench}. 
As a source of world knowledge and human values, we argue that LLMs enable foundation agents to align with the world model and human society created by languages, enhancing their reasoning and planning capabilities.

\section{Learning from Large-Scale Interactive Data}

Large-scale interactive data is an essential component for building foundation agents, akin to the significance of internet text and image data for foundation models in language and vision.
In this section, we first demonstrate how a potential foundation agent can be trained via offline IL when large, multi-modal and multi-task demonstrations are available.
We then discuss the potential use of data generation systems or real-world simulators for training foundation agents at scale.
Finally, we identify the constraints and alternatives to offline RL or IL methods in establishing foundation agents, as well as the limitations of real-world simulators realized by video generation models.

\subsection{Offline IL from Large Demonstrations}

Inspired by sequence modeling of RL problems, recent work attempt to train a generalist agent on extensive interactive demonstrations simply via offline IL.
For example, by imitating multi-modal demonstrations, Behavior Transformer \cite{shafiullah2022behavior} leverages a modified Transformer with action discretization and a multi-task action correction to capture the modes present in large behavior data.
Gato~\cite{reed2022generalist} unifies multi-modal and multi-task expert episodes into sequences through autoregressive sequence modeling, and is capable of diverse tasks and domains from robotic manipulation and Atari play to visual question answering.
Similarly, Unified-IO 2 \cite{lu2023unified} scales autoregressive multi-modal model across vision, language, audio, and action data, unifying different modalities into a shared semantic space.
With a single encoder-decoder Transformer, it learns varieties of skills through fine-tuning with prompts and augmentations.
However, collecting demonstrations via humans or online RL algorithms is costly and time-consuming, which is not applicable in real-world decision making.

\subsection{Potentials of Real-World Simulators}
Generative models as data generation systems~\cite{mandlekar2023mimicgen} or real-world simulators~\cite{yang2023learning,sora,bruce2024genie} open another path for training foundation agents through learning behaviors from generated data. 
For instance, RoboGen~\cite{wang2023robogen} realizes a propose-generate-learn cycle to scale up robot learning via generative simulation, of which policy learning algorithms are not restricted to IL but also involve motion planning or trajectory optimization.
Moreover, video generation models such as Sora~\cite{sora} and Genie~\cite{bruce2024genie} can be a general-purpose simulator of interactive decision making scenarios in physical and virtual worlds. 
These models, by unifying the representation of visual worlds~\cite{yang2024video}, generate diverse and open-ended training environments across tasks and domains for scaling up the foundation agent in a never-ending curriculum of new and generated worlds.
The foundation agent pretrained with these generated environments learns to reason about behaviors, handle environment stochasticity and uncertainty, and navigate competitive or cooperative multi-agent settings, thus enhancing adaptability and generalization in unseen scenarios.
This could be particularly valuable for domains where real-world interactive data is scarce or expensive to obtain, such as robotics and self-driving.

\begin{table*}[t]
    \centering
    \caption{Self-supervised pretraining objectives for decision making with Transformer. $P_\theta$ is the prediction of the pretrained model. $\tau_{0:t-1}$ denotes previous trajectories before timestep $t$. $s_t$ is the state or observation at timestep $t$ and $a_t$ the action at timestep $t$. $R_t$ is the return-to-go (target return for the rest of the input sequence) at timestep $t$, while $\hat{R_t}$ is the predicted future value and $r_t$ the future reward. $ G_{t+i}$ refers to a sub-goal or the ultimate goal at future timesteps, and $z$ denotes the encoded future trajectory in the same sequence length as the input trajectory.  
    Note that the conditional variables in each loss function are encoded by either a causal Transformer \protect\cite{brown2020language} for autoregressive modeling (top rows) or a bidirectional Transformer \protect\cite{devlin2018bert} for masked modeling (bottom rows).}
    \begin{tabular}{l|l}
    \toprule
    Pretraining Objective  & Loss Function  \\
    \hline
    Next action prediction &$-logP_\theta(a_t|\tau_{0:t-1},s_t)$ \\
    Reward-conditioned action prediction & $-logP_\theta(a_t|\tau_{0:t-1},s_t, R_t)$\\
    Future value and reward prediction & $-logP_\theta(a_t, \hat{R_t}, r_t|\tau_{0:t-1},s_t)$\\
    Goal-conditioned action prediction &$-logP_\theta(a_t|\tau_{0:t-1},s_t, G_{t+i})$ \\
    Future-conditioned action prediction &$-logP_\theta(a_t|\tau_{0:t-1},s_t, z)$ \\
    Forward dynamics prediction  &$-logP_\theta(s_t|\tau_{0:t-1})$ \\
    Inverse dynamics prediction  &$-logP_\theta(a_t|s_t, s_{t+1})$ \\
    \hline
    Random masking prediction &$-logP_\theta(masked(\tau)|unmasked(\tau))$ \\
    Reward-conditioned random mask prediction&
    $-logP_\theta(masked(\tau)|unmasked(\tau), R_0)$ \\
    Random masked hindsight prediction  &$-logP_\theta(masked(a)|unmasked(\tau), a_T)$ \\
    Random autoregressive mask prediction  &$-logP_\theta(masked(\tau, a_T)|unmasked(\tau))$ \\
    \bottomrule
    \end{tabular}
\label{tab:objective}
\end{table*}

\subsection{Discussion}
Despite the preliminary efforts, real-world interactive data scales far beyond internet text or visual data, making it impossible to solely rely on offline RL or IL to train foundation agents.
Alternative approaches, such as self-supervised (unsupervised) pretraining, can be utilized to harness large and unlabeled interactive data.
Moreover, current generalist agents remain small in model size compared to LLMs, but already consume significant computational resources even when trained on a single domain. We provide a summary of computational requirements of recently proposed generalist agents and a case we build in Appendix~\ref{append:computation}.

In addition, although we posit that universal video generation models (e.g., Sora and Genie) hold promise for enhancing the training of foundation agents, they are not the whole story.
Firstly, not all real-world decision making tasks can be adequately represented by video, such as situations when visual and textual information is not available like wireless communication and grid management. 
Secondly, the video generation task may not be a unified policy interface, since simply predicting next frame may not induce a reasonable agent behavior. 
Specifically, video generation objectives such as generating realistic video sequences given a fixed context, cannot be directly compatible with decision making objectives that optimize policies given feedback signals.
Thirdly, as the real world is much more complicated than simulators, the training of foundation agents should be grounded in simplified world such as a state formed as a result of an action. Then the agent can be trained on high-level results rather than low-level pixels and videos, which remains an open question for future work.

\section{Self-Supervised Pretraining and Adaptation}

Interactive data captures various aspects of information in the decision process, including state transitions, state-action causality and state or action values if rewards are provided. 
These decision-related knowledge should be learned during pretraining and transferred to downstream inference tasks so as to improve sample efficiency and generalization of agents.
Similar to foundation models in language and vision, we posit that the pretraining and adaptation pipeline can be considered as knowledge representation learning and reasoning of foundation agents.
In this section, we highlight strategies in self-supervised pretraining for decision making, and discuss the potential of pretraining and adaptation in building foundation agents.

\subsection{Self-Supervised Pretraining}

Self-supervised (unsupervised) pretraining for decision making allows foundation agents to learn without reward signals and encourages the agent to learn from suboptimal offline datasets.
This is particularly applicable when large, unlabeled data can be easily collected from internet or real-world simulators. 
Specifically, given a sequence of trajectories $\tau = \{(s_1,a_1,s_2,a_2,,...,s_T, a_T)\}^N$, self-supervised pretraining aims to learn a representation function $g:\mathbb{T}\in\mathbb{R}^d\to\mathbb{Z}\in\mathbb{R}^m$
$(m \ll d)$ to distill valuable knowledge from trajectory data for downstream inference. 
The knowledge can be temporal information about the same modality (e.g., $s_t \rightarrow s_{t+i}$), causal information between different modalities (e.g., $\pi(a|s)$), as well as dynamics $P(s^{\prime}|s,a)$ and reward $R(s,a)$ information.

Generally, there are two steps in pretraining foundation agents based on Transformer architecture.
The first step is to learn embeddings of trajectory data.
Specifically, the tokenization of trajectory sequences comprises three components:(1) \emph{trajectory encoding} that transforms raw trajectory inputs into a common representation space, (2) \emph{timestep encoding} that captures absolute or relative positional information, and (3) \emph{modality encoding} to disambiguate between different modalities in trajectories \cite{wu2023masked}.
Particularly, two levels of tokenization granularity have been studied: (1) discretization at the level of modalities \cite{sun2023smart,wu2023masked}, and (2) discretization at the level of dimensions \cite{reed2022generalist,boige2023pasta}.

The second step is devising self-supervised pretraining objectives to discover the underlying structure and semantics of trajectory data for knowledge representation learning.
Table~\ref{tab:objective} shows pretraining objectives for decision making in domains including control \cite{sun2023smart}, navigation \cite{carroll2022uni}, and game play \cite{lee2022multi}.
These objectives are mainly inspired by autoregressive \cite{radford2019language,brown2020language} or masked \cite{devlin2018bert,he2022masked} prediction in language and vision model pretraining.
The primary pretraining objective for decision making involves learning control information by predicting the next action in an autoregressive way. 
This objective is further modified by conditioning on different variables within trajectories, such as reward or value signals \cite{chen2021decision,lee2022multi}, next state (observation) information \cite{sun2023smart}, or latent future sub-trajectory information \cite{xie2023future}. 
As an alternative, random masking \cite{liu2022masked} learns the context of trajectory data by filling in missing information.
Various masking schemes conditioned on different RL components are designed.
For example, reward-conditioned random mask prediction \cite{carroll2022uni} recovers masked trajectory segments conditioned on the first-step return, while~\cite{sun2023smart} only recovers masked actions conditioned on the final-step action to capture global temporal relations for multi-step control.
Combining masked and autoregressive modeling, \cite{wu2023masked} constrains the last variable in trajectories to be masked to force the pretrained model to be causal at inference time.
Moreover, contrastive prediction objective \cite{stooke2021decoupling, schwarzer2021pretraining, yang2021representation, cai2023reprem} has been commonly used in self-supervised pretraining to learn state representation via a contrastive loss, which can benefit dynamics learning.

\subsection{Downstream Adaptation}

During adaptation, extensive decision-related knowledge acquired from pretraining is transferred to downstream tasks via fine-tuning or prompting. The knowledge transfer facilitates the optimization of learning objectives $f_\theta(z)$ in downstream inference, such as the value function $V(\pi)$ or $Q(s,a)$, policy, dynamics and reward functions.
The optimized objectives empowered by knowledge reasoning can finally improve sample efficiency and generalization compared to learning from scratch in traditional RL. 

Generally, there are two cases requiring fine-tuning: (1) when the pretraining data is a \emph{mix} of a small proportion of near-expert data and a large proportion of exploratory trajectories, and (2) when the pretraining objective \emph{significantly differs} from the inference objective of downstream decision making tasks. 
For instance, traditional RL aims to maximize cumulative rewards according to a specified reward function, whereas self-supervised pretraining tasks are usually reward-free. 
In such cases, RL algorithms are demanded for fine-tuning policies in online or offline settings \cite{yang2021representation,lee2022multi,sun2023smart,cai2023reprem,boige2023pasta}.

In addition to fine-tuning, prompting directly adapts the pretrained model to downstream inference without altering or introducing any parameters. It concerns: (1) \emph{aligning} pretraining objectives with downstream inference objectives, and (2) \emph{prompting} the model with interactive demonstrations~\cite{xu2022prompting,reed2022generalist,laskin2022context,wei2023imitation}.
For example, using a random masking pretraining objective with a variable mask ratio for different goals, \cite{liu2022masked} achieves zero-shot generalization to goal-reaching tasks.
This is attributed to the natural alignment of the masked pretraining objective with goal-reaching scenarios, where the model is required to recover masked actions based on remaining states.

\subsection{Discussion}

Previous attempts at self-supervised RL pretraining have been mostly limited to a single task \cite{yang2021representation}, or performing pretraining and fine-tuning within the same task \cite{schwarzer2021pretraining}.
This is not generic and flexible for adapting to various decision making tasks. 
Therefore, we argue for multi-modal and multi-task self-supervised pretraining in the evolution of foundation agents.
Particularly, self-supervised pretraining empowers foundation agents to acquire a nuanced understanding of large interactive data, laying a robust foundation for knowledge learning and reasoning in adaptation.
However, challenges arise in optimizing foundation agents, such as determining the optimal granularity and representation of trajectory data.
Furthermore, striking a balance between versatility and task specificity remains an ongoing challenge. 
Despite these problems, combining self-supervised pretraining and versatile adaptation strategies, underscores the potential of foundation agents in capable of robust performance across a broad array of decision making scenarios.

\section{Knowledge and Value Alignment via LLMs}

The broad real-world knowledge and human values embedded in LLMs pave the way for foundation agents with improved reasoning, generalization and interpretability.
In this section, we discuss the roles of LLMs in enhancing decision making realized by memory, planning and action modules.
We then outline some major challenges and potential solutions to align foundation agents with LLMs.

\subsection{Memory and Information Processing}

Foundation agents aligned with LLMs are often equipped with a memory module. The memory module functions as a repository, storing both task-specific information and past interaction history that enables agents to retrieve and refine relevant historical data for future planning. 
Similar as human memory \cite{izquierdo1999separate}, the agent's memory can be categorized into short-term and long-term components. 
Short-term memory encapsulates recent information within the context window, while long-term memory encompasses historical information stored in external storage.

After perceiving and processing environmental state, agents can decide what information to remember and what is crucial for current decision making that demands retrieval. For example, mastered skills \cite{wang2023voyager} are explicitly stored to prevent forgetting and retrieved as needed to enhance decision making efficiency. Further, through alignment with LLMs, foundation agents can refine their existing memories, generate new insights based on current knowledge, and assist future decision making \cite{park2023generative,shinn2023reflexion}.

\subsection{Planning with World Models}
The abundant world knowledge and human values embedded in LLMs enables LLMs to serve as world models \cite{hao2023reasoning}, simulating transition dynamics and reward functions \cite{ma2023eureka} for foundation agents to perform planning.
The agent then is able to reason about current observations and historical information, decomposing a complicated task into a sequence of strategic plans or sub-goals. 
Few-shot prompting \cite{zhao2021calibrate} can be one efficient means to utilize LLMs for enhancing agent's understanding and planning by providing high-quality demonstrations in short-term memory. For example, Chain-of-Thought (CoT) \cite{wei2022chain} decomposes tasks into logically coherent sub-goals, and Tree-of-Thought (ToT) \cite{yao2023tree} enables agents to generate potential solutions, allowing them to select the most reliable one. 
Furthermore, ReAct \cite{yao2022react} generates logically correct reasoning paths, improving the quality of generated plans by assessing correctness and logical coherence of previously generated actions.

Moreover, feedback from the environment \cite{yao2022react,huang2022inner,wang2023voyager,zhu2023ghost,wang2023describe}, LLM critics \cite{shinn2023reflexion,madaan2023self,chen2023interact}, or humans \cite{huang2022inner} can also be leveraged to improve the planning of foundation agents. These feedback enables agents to adjust behaviors and plans, addressing obstacles encountered during execution and thus improving task fulfillment.

\subsection{Action and Decision Making}

For ultimate action execution and interaction with real or simulated environments, an action module is required for foundation agents to interpret plans from the planning module, translate subgoals into executable actions, and combine atomic actions to form more complex structured actions. 

Specifically, the action module specifies action goals and available actions. 
Action goals articulate intended objectives, such as movement \cite{zhu2023ghost}, communication \cite{zhang2023building}, and code generation \cite{wang2023voyager}. Clear goal descriptions enable agents to comprehend and generate reasonable actions based on observations. 
Available actions, on the other hand, define the agent's action space, which can be expanded by adding recently acquired skills into memory \cite{wang2023voyager,zhu2023ghost}. 
Additionally, equipped with LLMs, foundation agents can possess the ability to call external APIs or tools for problem solving. 
HuggingGPT \cite{shen2023hugginggpt} exemplifies this by using LLMs as the controller to manage existing AI models from the internet via APIs to execute each sub-task.
Further, by connecting to external databases \cite{hu2023chatdb,zhu2023ghost}, the agent can acquire extra task-related knowledge via external tools and plugins created by domain experts \cite{ge2023openagi} for reliable and creative decision making.

\subsection{Discussion}
The integration of LLMs into agent learning offers advantages over conventional RL and IL methods. Firstly, LLMs require significantly fewer training samples \cite{guo2023suspicion}, as they leverage pretraining on internet-scale text, acquiring extensive real-world knowledge and demonstrating strong generalization abilities. 
Secondly, state-of-the-art LLMs are inherently human-aligned, enabling them to make decisions consistent with human preferences and values \cite{zhang2023building}. 
Thirdly, by providing detailed reasoning for actions during planning, the interpretability of foundation agents aligned with LLMs addresses the challenge of deploying black-box AI models \cite{dovsilovic2018explainable} in real-world decision making. 

However, a substantial challenge for aligning foundation agents with LLMs lies in the hallucination of LLMs . Hallucination refers to the phenomenon where the language model generates false content that is misaligned with real-world knowledge \cite{zhang2023siren}. Since hallucination detection still remains an open problem, aligning foundation agents with LLMs amplifies the risk of inexplicable abnormal behaviors. Given the broad range of tasks assigned to foundation agents, particularly those that are safety-critical, the hallucination in LLMs could result in serious consequences during decision making. Therefore, it is crucial to remain vigilant about the potential risks of foundation agents aligned with LLMs, and propose solutions to this issue for a safe and responsible deployment in real-world decision making.

\section{Trends for Foundation Agents}
Derived from the formulation and challenges for foundation agents, some critical research questions remain to be solved. In this section, we propose these issues and potential solutions, and discuss the trends for foundation agents with examples in recent work as well as a case study in real-world decision making scenarios.

\subsection{A Unified or Compositional Foundation Agent}
\label{sec:morphology}

One central challenge in building foundation agents stems from the substantial diversity among decision making tasks, involving variations in state and action spaces, environment dynamics, and reward functions. 
In contrast, the unification of tasks in language is facilitated by the text-to-text generation paradigm \cite{brown2020language,gpt4}. 
To structure diverse environments and tasks, some attempts have been made along two primary directions: (1) the pursuit of a \emph{unified} foundation model \cite{reed2022generalist,brohan2022rt}, and (2) the exploration of \emph{compositions} involving multiple existing foundation models \cite{zeng2022socratic,ajay2023compositional} for decision making.

To establish a unified foundation model for decision making, a core requirement is the unified representation of RL components. This entails encoding states, actions, and rewards from diverse environments and tasks into standardized tokens through sequence modeling \cite{reed2022generalist,Janner2021OfflineRL,chen2021decision}. The subsequent step involves transforming these tokens into a consistent data modality, such as text descriptions \cite{zhu2023ghost}, generated videos \cite{du2023learning}, code \cite{wang2023voyager}, or text-image pairs \cite{brohan2023rt}.
However, due to the multi-modal and heterogeneous characteristics of interactive data, uncertainties persist regarding the efficacy of tokenization methods in compressing raw trajectories into compact tokens.
In addition, whether a single data modality can comprehensively represent interactive data and effectively convey its underlying information and knowledge warrants further investigation.

While a unified foundation model might offer a comprehensive solution, it could also lead to increased complexity and challenges in interpretability. 
Instead, we consider whether integrating existing foundations models (e.g., large language or vision models) with moderate decision models is sufficient to address most (if not all) of decision making tasks.
The integration of existing foundation models allows for leveraging their domain-specific strengths but may introduce issues in harmonizing different modalities and functionalities.
Additionally, a compositional foundation agent could inherit both merits and demerits from other foundation models. 
Therefore, striking a balance between these approaches requires careful evaluation of specific requirements and constraints of target tasks, offering an intriguing avenue for future research in foundation agents.

\subsection{Policy Optimization with Foundation Agents}

In traditional RL, theoretical guarantees underpin algorithmic advancements, such as the Bellman optimality update in temporal difference learning and dynamic programming. Recent theoretical analyses have shed light on the delusions in sequence models for interaction and control \cite{ortega2021shaking}, yet the optimization of a generalist policy with foundation agents still lacks a solid theoretical foundation.

Establishing theoretical foundations for policy optimization with foundation agents is a complex but crucial endeavor.
Firstly, policies generated by foundation agents will directly affect changes in real-world applications, requiring rigorous theoretical guarantees to ensure their effectiveness, optimality, safety and robustness. Meanwhile, the theoretical foundations for policy optimization can in turn help to understand the principles underlying the agent's behavior and performance, thereby enabling the development of algorithms and methodologies for training foundation agents.
Specifically, we posit that the theoretical foundations require an interdisciplinary view (e.g., control theory, RL, and optimization), and we outline some ideas here.

\begin{itemize}[leftmargin=*,nolistsep]
    \item \textbf{Define Pretraining and Task-Specific Objectives}: Unlike traditional RL objectives that focus on maximizing cumulative rewards, the optimization objectives for foundation agents involve both pretraining and task-specific objectives. Pretraining objectives address the unified representation of variables involved in decision process and the unified policy interface across tasks and domains. Task-specific objectives may be employed in fine-tuning phase, which have been established in many control, planning, and RL algorithms~\cite{yang2023learning,wang2023robogen}.
    \item \textbf{Understand the Interplay between Pretraining and Task-Specific Optimization}: Establish a mathematical framework to investigate how the pretraining phase shapes the optimization landscape for task-specific policy optimization, and how different adaptation algorithms interact with the previously acquired knowledge. This could involve probing convergence properties, sample efficiency, and generalization bounds of foundation agents. 
    \item \textbf{Extend Existing Theoretical Frameworks}: Adapt and extend existing theoretical frameworks from RL, control theory and optimization to accommodate the peculiarities of foundation agents. This may involve relaxing assumptions made in traditional RL and developing new notions of optimality for generalist policies. For example, (1) \emph{extend the MDP framework} with generalized state spaces to incorporate multi-modal inputs and extended action spaces spanning different domains (e.g., physical movements and linguistic responses), (2) leverage the probabilistic representations and principled reasoning offered by \emph{the control as inference} framework~\cite{todorov2008general} to derive optimal control policies and analyze their properties, and (3) ground policy optimization for foundation agents in \emph{information-theoretic} principles, such as mutual information or entropy, to capture the agent's ability to influence the environment and gather informative experiences. 
\end{itemize}

\subsection{Open-Ended Tasks for Foundation Agents}

In the evolution of foundation agents, a notable trend is the shift towards learning from open-ended tasks at scale \cite{team2021open,fan2022minedojo}. Traditionally, RL agents are confined to a single and individual task \cite{silver2017mastering,schwarzer2023bigger}, limiting their applicability to massive dynamic scenarios. Contemporary approaches instead prioritize the scalability of generative models, especially via LLMs to effectively handle the complexities inherent in open-ended tasks~\cite{wang2023voyager,zhu2023ghost}.
The open-ended agents, endowed with continuously learning capabilities, interact with environments or external tools \cite{shen2023hugginggpt,hu2023chatdb,ge2023openagi}, iteratively refining their decision process. This paradigm aligns seamlessly with RL principles, where agents acquire optimal strategies through iterative learning.

Despite the promising advances in the development of open-ended agents, several challenges persist particularly for open-ended tasks. Importantly, open-ended tasks do not have a predefined goal or endpoint, push the boundaries of agent adaptability, generalization, and continual learning, and require flexible adaptation, creativity, and the ability to discover new goals or objectives. Inspired by these features, we discuss the issues associated with foundation agent learning in open-ended tasks, along with insights into how these issues can be addressed.

\begin{itemize}[leftmargin=*,nolistsep]
    \item \textbf{Continual Learning and Adaptation}: The dynamic nature of open-ended tasks poses challenges in \emph{model adaptability without catastrophic forgetting} of acquired skills and knowledge. Agents need to handle an unlimited variety of open-ended goals with training objectives and task distributions dynamically changing. Two lines of recent work attempt to address the issue by (1) leveraging pretrained language or vision model to convert world knowledge into actionable insights~\cite{ahn2022can,driess2023palm}, and (2) pretraining (or meta-learning) large transformer models from scratch on multi-tasks~\cite{team2021open,fan2022minedojo,bauer2023human}. However, these studies require mostly human-defined goals as input. Instead, \cite{wang2023toward} formulates goal as energy function and learns without goal-conditioning via diffusion models. Nevertheless, it is also limited by human-designed functions for goals which could be replaced by neural networks in future work.
    \item \textbf{Unboundedness and Novelty Detection}: In open-ended tasks, the environment might be \emph{infinite or continually evolving}, making it difficult to exhaustively explore or anticipate all possible states and outcomes. Agents need to deal with \emph{novelty and unexpected situations} throughout their lifetime. Current research attempts to address this issue through novelty difficulty measures~\cite{pinto2023difficulty}, hierarchical RL~\cite{vezhnevets2017feudal} and self-initiated open world learning framework~\cite{liu2023ai}, yet it still remains an open question.
    \item \textbf{Curriculum and Autotelic Learning}: Agents learning in open-ended tasks have to learn and explore \emph{without explicit external rewards}. Existing studies apply procedural environment generation to create diverse and increasingly complex environments~\cite{bauer2023human}, and intrinsic motivation~\cite{forestier2022intrinsically} and curiosity-driven learning to encourage exploration. Future work can consider automatically generate curriculum via LLMs~\cite{wang2023voyager}, universal video generation models~\cite{sora,bruce2024genie} or multi-modal models.
    \item \textbf{Creativity and Innovative Solutions}: Generating \emph{novel and valuable solutions} is a hallmark of open-ended learning. This necessitates agents that can go beyond mere imitation or optimization, possibly by combining known concepts in unique ways or inventing entirely new ones. Potential solutions include (1) Quality-Diversity algorithms~\cite{grillotti2022unsupervised} to produce a diverse set of high-performing solutions, (2) neuroevolution methods~\cite{lehman2011abandoning} to produce innovative solutions, and (3) deep generative models to generate novel and sometimes unexpectedly creative outputs.
\end{itemize}

\begin{figure}
    \centering
    \includegraphics[scale=0.3]{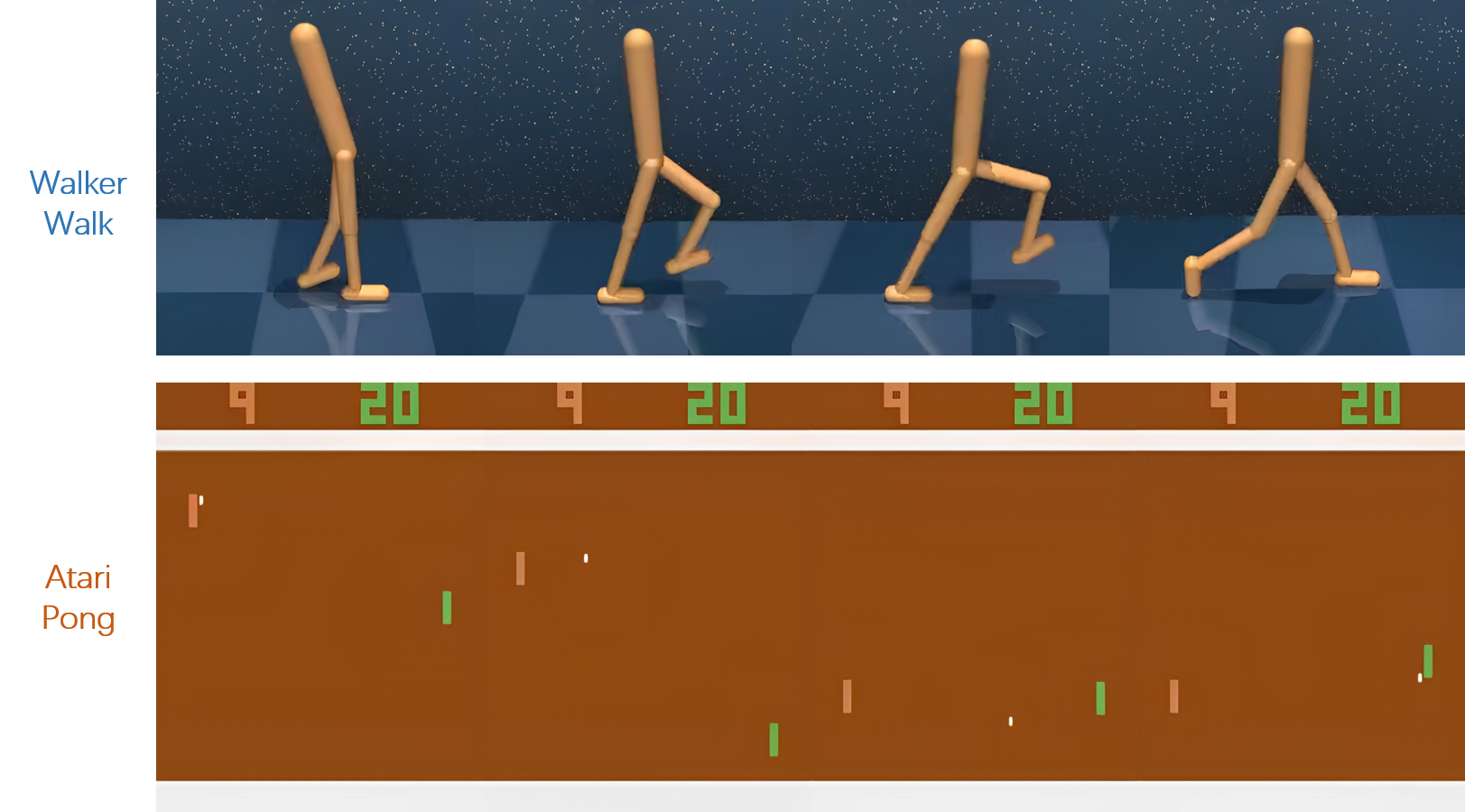}
    \caption{Visualized fine-tuning performance of our pretrained agent on unseen task and game. Frames are selected from the video recorded in the last evaluation episode. Total number of evaluation episodes is 50. Results are from one training seed.}
    \label{fig:case-study}
\end{figure}

\subsection{Case Studies in Real-World Decision Making}

\textbf{Autonomous Control.}
The progress in robotics continually enhances productivity and resource efficiency.
Recently, robots with general capabilities have been developed towards foundation agents.
For example. RT-1~\cite{brohan2022rt} is a generalist and language-conditioned robotic agent, exhibiting zero-shot generalization to new tasks, environments and long-horizon scenarios by scaling on a large set of manipulation demonstrations~\cite{padalkar2023open}.
RT-2~\cite{brohan2023rt} further improves generalization and reasoning in response to user commands by co-fine-tuning pretrained vision-language models on demonstrations and internet vision-language tasks.
Moreover, equipped with real-time human feedback, robots can also improve their high-level decision making iteratively in long-horizon tasks~\cite{shi2024yell}.
We also conduct a case study in robotic locomotion and game play by jointly pretraining on 5 tasks in the DeepMind Control (DMC) suite~\cite{tassa2018deepmind} and 5 games in Atari video play~\cite{bellemare2013arcade} through autoregressive modeling. Visualization of the fine-tuning performance on a new locomotion task and a new game is shown in Figure~\ref{fig:case-study}. Details of our case study can be found in Appendix~\ref{append:case}.

In addition to robotics and game play, self-driving is another critical domain where foundation agents have major impact opportunities. As a paradigm shift from previous perception-prediction-planning framework, LLMs are exploited to integrate common-sense knowledge and human cognitive abilities into autonomous driving~\cite{wen2023dilu,mao2023language}, and enable multiple vehicles to realize collaboration over time \cite{hu2024agentscodriver}. Despite these trends for foundation agents in autonomous control, safety, robustness and reliability of decisions made by agents requires rigorous evaluation and assurance in real-world deployment.

\textbf{Healthcare.}
Human-Agent collaborative decision making in healthcare has the opportunity to enhance diagnostic accuracy, streamline operations, and provide personalized treatment options.
Current research in foundation agents for healthcare is rare but has shown effectiveness in assessing patient's state in ICU (intensive care unit) rooms~\cite{durante2024interactive} and designing treatment plans through discussions among LLMs as different medical experts~\cite{tang2023medagents}. 
However, agents in healthcare still raise high uncertainty and distrust from clinicians.
Inspired by drawing biomedical literature evidence via LLM~\cite{yang2023harnessing} or involving human experts in the intermediate stages of medical decision making~\cite{zhang2024rethinking}, real-world healthcare can be realized through consistent collaboration between humans and foundation agents in the future.

\textbf{Science.}
Scientific discovery and research can be revolutionized by foundation agents with significantly accelerated data analysis and experimentation processes, leading to faster and more accurate insights.
AlphaFold 3~\cite{abramson2024accurate} has demonstrated the feasibility of high accuracy modelling across biomolecular space within a single unified model, consistent with the characteristics of foundation agents that have a unified representation and policy interface across state-action spaces. Aligned with LLMs and empowered by external tools, chemistry agents~\cite{boiko2023autonomous,m2024augmenting} show their potential for accelerating research and discovery.
They can be applied in organic synthesis, reaction optimization, drug discovery, and materials design with capabilities of (semi-) autonomous hypothesis generation, experimental design and execution.
However, the deployment of foundation agents in scientific research poses potential biases in data interpretation, unintended experimental errors, and ethical concerns regarding the transparency and accountability of findings, requiring rigorous oversight, comprehensive validation and ethical guidelines.

\section{Conclusion}

We take the position that foundation agents hold the potential to alter the landscape of agent learning for decision making, akin to the revolutionary impact of foundation models in language and vision.
We support this position by formulating the notion of foundations agents and specifying their roadmap and key challenges.
Demonstrated by recent work and a case study, we show that the enhanced perception, adaptation, and reasoning abilities of agents not only address limitations of conventional RL, but also hold the key to unleash the full potential of foundation agents in real-world decision making.
However, due to the intrinsic complexity of decision making, existing work towards foundation agents is still at a starting point.
Future directions involve but are not limited to learning a unified representation of diverse environments and tasks, establishing theoretical foundations for a unified policy interface and learning from open-ended tasks at scale. Bridging these gaps will pave the way for more robust and versatile foundation agents, contributing to the evolution of artificial general intelligence.

\section*{Acknowledgements}
This work is supported in part by the Strategic Priority Research Program of Chinese Academy of Sciences (Grant No.XDA27010103) and the Youth Innovation Promotion Association CAS.

\section*{Impact Statement}
The development and deployment of foundation agents carry profound implications for society, shaping the landscape of AI and decision-making systems. The potential broader impact spans ethical considerations, societal consequences, and the democratization of AI. Ethically, foundation agents could contribute to more transparent and accountable decision-making processes, emphasizing the importance of fairness, interpretability, and adherence to ethical standards. However, it also raises concerns about privacy, bias, and unintended consequences, necessitating robust ethical frameworks and governance. Societally, widespread adoption of foundation agents has the potential to revolutionize various sectors, from autonomous control and healthcare to scientific research and beyond, 
enhancing efficiency and accessibility. Yet, it requires careful consideration of its impact on employment, socioeconomic disparities, and digital divide issues. Striking a balance between innovation and responsible deployment is crucial to harness the full societal potential of foundation agents while mitigating unintended negative consequences. Future developments in this space should be guided by a commitment to ethical principles, social responsibility, and the promotion of positive societal transformations.

\nocite{langley00}

\bibliography{main}
\bibliographystyle{icml2024}

\newpage
\appendix
\onecolumn
\section{Distinctions between Foundation Agents, Task-Specific Agents and LLM-based Agents}
\label{append:distinction}

Compared to conventional task-specific agents developed by RL or IL approaches, the foundation agent has the capabilities of processing multi-modalities, adapting to new tasks and domains, and few- or zero-shot generalization with prompting.
Compared to LLM-based agents that mainly carry out linguistic or knowledge-based tasks, the foundation agent does physical planning, optimizes decisions, and plays two- or multi-player games in physical and virtual worlds. An LLM may be part of a foundation agent to enable the agent to align with the world model and human society created by languages. However, as stated in Section~\ref{sec:morphology}, the morphology of a foundation agent is not necessarily restricted to a composition of several existing foundation models. Instead, an LLM-based agent in current research~\cite{wang2024survey} places LLMs at the core of decision making.

\section{Computational Requirements of Potentially Foundation Agents}
\label{append:computation}

Table~\ref{tab:compu} shows the model size and computational requirements of some generalist agents and our case study towards foundation agents.
Recent works~\cite{lee2022multi,brohan2022rt,brohan2023rt,wei2023imitation,chebotar2023q,durante2024interactive} demonstrate that these agents are able to continue to scale to larger dataset sizes and model sizes with performance improvement. Therefore, training a foundation agent may require much more computational resources than expected, and scaling issues involving hardware acceleration, parallel processing, and model complexity and optimization need to be considered.
In future work, the scaling law concerning the data and model size for training foundation agents needs to be addressed.

\begin{table}[h!]
    \centering
    \caption{Model size and computational requirements of current potential foundation agents and our case study. Results except for the last row are from original papers. Note that these agents are still far from the truly foundation agent as defined in our paper, and we only include those whose computational requirements are explicitly described in their original papers.}
    \begin{tabular}{l|l|l}
    \toprule
       Model  & Params & Computational Requirements\\
       \hline
       Gato~\cite{reed2022generalist} & 1.18B & Training 4 days on 256 TPUv3 \\
       \hline
        Multi-Game Decision Transformer~\cite{lee2022multi} & 200M & Preraining 8 days on 64 TPUv4\\
        \hline
        \multirow{2}*{DualMind~\cite{wei2023imitation}} & 175M (pretraining) & 40h in Phase I on 40 V100\\
        &51.1M (IL from prompts) & 12h in Phase II on 16 V100 \\
        \hline
        Interactive Agent Foundation Model~\cite{durante2024interactive} & 277M & Pretraining 175h on 16 V100 \\
        \hline
        Our case & 25M & Pretraining 8h on 4 RTX4090 \\
    \bottomrule
    \end{tabular}
    \label{tab:compu}
\end{table}

\section{Case Study in DMC and Atari}
\label{append:case}
\subsection{Environments and Datasets}
\textbf{DeepMind Control Suite.}
The DeepMind Control Suite~\cite{tassa2018deepmind} is a set of physics-based simulation environments, containing continuous control tasks with image observations spanning diverse domains and tasks. Different domains are associated with different state-actions spaces and each task is associated with a particular MDP within each domain.
We select 5 tasks for pretraining and one unseen task for fine-tuning as shown in Table~\ref{tab:dmc}.
For each task, we collect trajectories using pixels from the training progress of a SAC~\cite{haarnoja2018soft} agent for 1M steps to include both expert and non-expert data for pretraining.
For fine-tuning, we randomly sample 10\% trajectories from the full replay buffer of the SAC agent with diverse return distribution.

\begin{table}[h!]
    \caption{Domains and tasks used in pretraining and fine-tuning on DMC.}
    \centering
    \begin{tabular}{c|c|c}
    \toprule
      Phase   &  Domain & Task \\
      \hline
       \multirow{5}*{Pretraining} & cartpole   & swingup\\
       ~ & hopper  & hop  \\
       ~ & cheetah  & run  \\
       ~ & walker  & stand  \\
       ~ & walker  & run  \\
      \hline
      Fine-tuning &  walker  &  walk\\
      \bottomrule
    \end{tabular}
    \label{tab:dmc}
\end{table}

\textbf{Atari.}
The Atari benchmark~\cite{bellemare2013arcade} is a collection of video game environments with distinctly different dynamics, rewards, and agent embodiments.
We select 5 games for pretraining and one unseen game for fine-tuning as shown in Table~\ref{tab:atari}.
For each game, we collect trajectories using pixels from the training progress of a DQN~\cite{mnih2015human} agent for 1M steps to include both expert and non-expert data for pretraining.
For fine-tuning, we randomly sample 10\% trajectories from the full replay buffer of the DQN agent with diverse return distribution.

\begin{table}[h!]
\caption{Atari games used in pretraining and fine-tuning.}
    \centering
    \begin{tabular}{c|c}
    \toprule
     Phase    &  Game\\
     \hline
     \multirow{5}*{Pretraining} & Asterix\\
       ~ & Breakout  \\
       ~ & Gopher  \\
       ~ & Qbert  \\
       ~ & Seaquest  \\
       \hline
    Fine-tuning & Pong\\
    \bottomrule
    \end{tabular}
    \label{tab:atari}
\end{table}

\subsection{Implementation Details}
\textbf{Model Architecture.}
The implementation of our agent is based on GPT-2~\cite{brown2020language} with configurations summarized in Table~\ref{tab:hyperparam}.
The input for pretraining consists of a sequence of image observations and actions, while during fine-tuning returns are also added as input for RL.
The observation tokenizer is a single ResNet~\cite{he2016deep} block to obtain image patch token embeddings following~\cite{reed2022generalist}.
Action tokenizer and return tokenizer are single-layer linear prediction heads.
We follow the tokenization scheme and add patch position encodings and local observation position encodings to token embeddings as introduced in~\cite{reed2022generalist}.
Similar to~\cite{reed2022generalist,lee2022multi}, we model the data autoregressively using a sequential causal
attention masking but allow observation tokens within the same
timestep to access each other.

\begin{table}[h!]
\caption{Transformer hyperparameters of our agent.}
    \centering
    \begin{tabular}{c|c}
    \toprule
    Hyperparameter     &  Model\\
    \hline
    Transformer blocks     & 8\\
    Attention heads & 16\\
    Layer width & 512\\
    Feedforward hidden size & 2048\\
    \hline
    Shared embedding & True\\
    Layer normalization & Pre-norm\\
    Activation function & GeGLU \\
    \bottomrule
    \end{tabular}
    \label{tab:hyperparam}
\end{table}

\textbf{Pretraining and Fine-tuning.}
We jointly pretrain our agent on 5 DMC tasks and 5 Atari games using the pretraining objective of next action prediction $-logP_\theta(a_t|\tau_{0:t-1},o_t)$ via standard mean-squared error loss for continuous actions or cross-entropy loss for discrete actions.
We use the AdamW optimizer~\cite{loshchilov2017decoupled} with parameters $\beta_1=0.9$, $\beta_2=0.95$, and $\epsilon=1e-8$ for 1M steps.
For pretraining, the learning rate is originally set to be 1e-5 with linear warm-up and cosine schedule decay.
We use batch size of 512 and weight decay of 0.1 for pretraining. 
For fine-tuning, the learning rate keeps constant of 1e-4 with batch size 256 and dropout rate 0.1.
During fine-tuning, we model and predict returns and rewards in addition to actions using the learning objective $-logP_\phi(R_t,a_t,r_t|\tau_{0:t-1},o_t)$, and perform expert action inference as introduced in~\cite{lee2022multi}.
We use context length $L=50$ for both pretraining and fine-tuning\footnote{Our code is based on \url{https://github.com/microsoft/smart}.}.
The size of our pretrained model and corresponding computational requirements can be referred to Table~\ref{tab:compu}.
Since this is simply a case study of potential foundation agent in applications of robotics and game play, we only provide the visualization of fine-tuning performance in Figure~\ref{fig:case-study} to demonstrate its feasibility and effectiveness in decision making scenarios.
We do not report other experimental results and compare with baselines as done in original research.
In addition, the case study only involves the first two steps in the roadmap of foundation agents as specified in Figure~\ref{fig:perspectives}. We consider aligning the pretrained model in this case study with LLMs in future work.

\end{document}